# Feature Squeezing:
## Detecting Adversarial Examples in Deep Neural Networks

Weilin Xu, David Evans, Yanjun Qi
*University of Virginia*
evadeML.org

*Abstract*—Although deep neural networks (DNNs) have achieved great success in many tasks, they can often be fooled by *adversarial examples* that are generated by adding small but purposeful distortions to natural examples. Previous studies to defend against adversarial examples mostly focused on refining the DNN models, but have either shown limited success or required expensive computation. We propose a new strategy, *feature squeezing*, that can be used to harden DNN models by detecting adversarial examples. Feature squeezing reduces the search space available to an adversary by coalescing samples that correspond to many different feature vectors in the original space into a single sample. By comparing a DNN model's prediction on the original input with that on squeezed inputs, feature squeezing detects adversarial examples with high accuracy and few false positives. This paper explores two feature squeezing methods: reducing the color bit depth of each pixel and spatial smoothing. These simple strategies are inexpensive and complementary to other defenses, and can be combined in a joint detection framework to achieve high detection rates against state-of-the-art attacks.

## 1. Introduction

Deep Neural Networks (DNNs) perform exceptionally well on many artificial intelligence tasks, including security-sensitive applications like malware classification [26] and face recognition [37]. Unlike when machine learning is used in other fields, security applications may involve sophisticated adversaries responding to the deployed systems. Recent studies have shown that attackers can force deep learning object classification models to misclassify images by making imperceptible modifications to pixel values. The maliciously generated inputs are called "adversarial examples" [10, 40] and are normally crafted using an optimization procedure to search for small, but effective, artificial perturbations.

The goal of this work is to harden DNN systems against adversarial examples by detecting adversarial inputs. Detecting an attempted attack may be as important as predicting correct outputs. When running locally, a classifier that can detect adversarial inputs may alert its users or take fail-safe actions (e.g., a fully autonomous drone returns to its base) when it spots adversarial inputs. For an on-line classifier whose model is being used (and possibly updated) through API calls from external clients, the ability to detect adversarial examples may enable the operator to identify malicious clients and exclude their inputs. Another reason that detecting adversarial examples is important is because even with the strongest defenses, adversaries will occasionally be able to get lucky and find an adversarial input. For asymmetrical security applications like malware detection, the adversary may only need to find a single example that preserves the desired malicious behavior but is classified as benign to launch a successful attack. This seems like a hopeless situation for an on-line classifier operator, but the game changes if the operator can detect even unsuccessful attempts during an adversary's search process.

Most previous work on hardening DNN systems, including *adversarial training* and *gradient masking* (details in Section 2-C), focused on modifying the DNN models themselves. In contrast, our work focuses on finding simple and low-cost defensive strategies that alter the input samples but leave the model unchanged. Section 2-D describes a few other recent proposals for detecting adversarial examples.

Our approach, which we call *feature squeezing*, is driven by the observation that the feature input spaces are often unnecessarily large, and this vast input space provides extensive opportunities for an adversary to construct adversarial examples. Our strategy is to reduce the degrees of freedom available to an adversary by "squeezing" out unnecessary input features. The key idea is to compare the model's prediction on the original sample with its prediction on the sample after squeezing, as depicted in Figure 1. If the original and squeezed inputs produce substantially different outputs from the model, the input is likely to be adversarial. By comparing the difference between predictions with a selected threshold value, our system outputs the correct prediction for legitimate examples and rejects adversarial inputs.

Although feature squeezing generalizes to other domains,

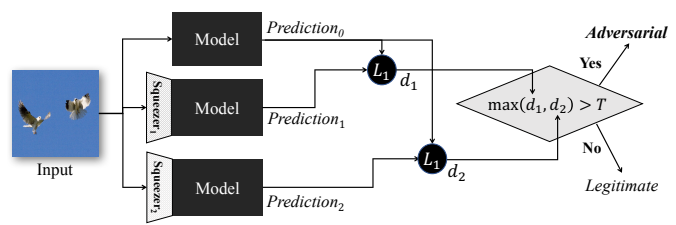

Fig. 1: **Feature-squeezing framework for detecting adversarial examples.** The model is evaluated on both the original input and the input after being pre-processed by feature squeezers. If the difference between the model's prediction on a squeezed input and its prediction on the original input exceeds a threshold level, the input is identified to be adversarial.

here we focus on image classification. Because it is the domain where adversarial examples have been most extensively studied. We explore two simple methods for squeezing features of images: reducing the color depth of each pixel in an image and using spatial smoothing to reduce the differences among individual pixels. We demonstrate that feature squeezing significantly enhances the robustness of a model by predicting correct labels of non-adaptive adversarial examples, while preserving the accuracy on legitimate inputs (Section 4), thus enabling an accurate detector for static adversarial examples (Section 5). Feature squeezing appears to be both more accurate and general, and less expensive, than previous methods, though the robustness against adaptive adversary needs further investigation in the future work.

**Contributions.** Our key contribution is introducing and evaluating feature squeezing as a technique for detecting adversarial examples. We show how the general detection framework (Figure 1) can be instantiated to accurately detect adversarial examples generated by several state-of-the-art methods. We study two instances of feature squeezing: reducing color bit depth (Section 3-A) and both local and non-local spatial smoothing (Section 3-B). We report on experiments that show feature squeezing helps DNN models predict correct classification on adversarial examples generated by eleven different and state-of-the-art attacks mounted without knowledge of the defense (Section 4). Feature squeezing is complementary to other adversarial defenses since it does not change the underlying model, and can readily be composed with other defenses such as adversarial training (Section 4-B).

Section 5 explains how we use feature squeezing for detecting static adversarial inputs, combining multiple squeezers in a joint detection framework. Our experiments show that joint-detection can successfully detect adversarial examples from eleven static attacks at the detection rates of 98% on MNIST and 85% on CIFAR-10 and ImageNet, with low (around 5%) false positive rates. Although we cannot guarantee an adaptive attacker cannot succeed against a particular feature squeezing configuration, our results show it is effective against state-of-the-art static methods, and it considerably complicates the task of an adaptive adversary even with full knowledge of the model and defense (Section 5-D).

## 2. Background

This section provides a brief introduction to neural networks, adversarial examples, and previous defenses.

### A. Neural Networks

Deep Neural Networks (DNNs) can efficiently learn highly-accurate models from large corpora of training samples in many domains [18, 26]. Convolutional Neural Networks (CNNs), popularized by LeCun et al. [20], perform exceptionally well on image classification. A deep CNN can be written as a function $g : X \rightarrow Y$, where $X$ represents the input space and $Y$ is the output space representing a categorical set. For a sample, $\mathbf{x} \in X$, $g(\mathbf{x}) = f_L(f_{L-1}(\ldots((f_1(\mathbf{x}))))$. Each $f_i$ represents a layer. The last output layer, $f_L$, creates the mapping from a hidden space to the output space (class labels) through a softmax function that outputs a vector of real numbers in the range [0, 1] that add up to 1. We can treat the output of softmax function as the probability distribution of input $\mathbf{x}$ over $C$ different possible output classes.

A training set contains $N$ labeled inputs where the $i^{\text{th}}$ input is denoted $(\mathbf{x}_i, y_i)$. When training a deep model, parameters related to each layer are randomly initialized, and input samples, $(\mathbf{x}_i, y_i)$, are fed through the network. The output of this network is a prediction $g(\mathbf{x}_i)$ associated with the $i^{\text{th}}$ sample. To train the DNN, the difference between prediction output, $g(\mathbf{x}_i)$, and its true label, $y_i$, usually modeled with a loss function $J(g(\mathbf{x}_i), y_i)$, is pushed backward into the network using a back-propagation algorithm to update DNN parameters.

### B. Generating Adversarial Examples

An adversarial example is an input crafted by an adversary with the goal of producing an incorrect output from a target classifier. Since ground truth, at least for image classification tasks, is based on human perception which is hard to model or test, research in adversarial examples typically defines an adversarial example as a misclassified sample $\mathbf{x}'$ generated by perturbing a correctly-classified sample $\mathbf{x}$ (the *seed* example) by some limited amount.

Adversarial examples can be *targeted*, in which case the adversary's goal is for $\mathbf{x}'$ to be classified as a particular class $t$, or *untargeted*, in which case the adversary's goal is just for $\mathbf{x}'$ to be classified as any class other than its correct class. More formally, given $\mathbf{x} \in X$ and $g(\cdot)$, the goal of an targeted adversary with target $t \in Y$ is to find an $\mathbf{x}' \in X$ such that

$$g(\mathbf{x}') = t \wedge \Delta(\mathbf{x}, \mathbf{x}') \leq \epsilon \tag{1}$$

where $\Delta(\mathbf{x}, \mathbf{x}')$ represents the difference between input $\mathbf{x}$ and $\mathbf{x}'$. An untargeted adversary seeks to find an $\mathbf{x}' \in X$ such that

$$g(\mathbf{x}') \neq g(\mathbf{x}) \wedge \Delta(\mathbf{x}, \mathbf{x}') \leq \epsilon. \tag{2}$$

The strength of the adversary, $\epsilon$, limits the permissible transformations. The distance metric, $\Delta(\cdot)$, and the adversarial strength threshold, $\epsilon$, are meant to model how close an adversarial example needs to be to the original to "fool" a human observer.

Several techniques have been proposed to find adversarial examples. Szegedy et al. [40] first observed that DNN models are vulnerable to adversarial perturbation and used the Limited-memory Broyden-Fletcher-Goldfarb-Shanno (L-BFGS) algorithm to find adversarial examples. Their study also found that adversarial perturbations generated from one DNN model can also force other DNN models to produce incorrect outputs. Subsequent papers have explored other strategies to generate adversarial manipulations, including using the linear assumption behind a model [10, 28], saliency maps [34], and evolutionary algorithms [29].

Equations (1) and (2) suggest two different parameters for categorizing methods for finding adversarial examples: whether they are targeted or untargeted, and the choice of $\Delta(\cdot)$, which is typically an $L_p$-norm distance metric. Popular adversarial methods used the following three norms for $\Delta(\cdot)$:

- $L_\infty$: $\|\mathbf{z}\|_\infty = \max_i |z_i|$.
  The $L_\infty$ norm measures the maximum change in any dimension. This means an $L_\infty$ attack is limited by the maximum change it can make to each pixel, but can alter all the pixels in the image by up to that amount.

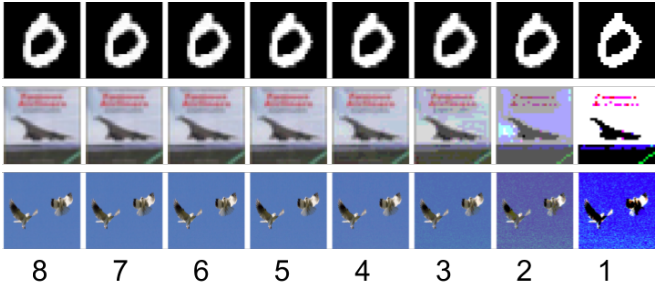

Fig. 2: Image examples with bit depth reduction. The first column shows images from MNIST, CIFAR-10 and ImageNet, respectively. Other columns show squeezed versions at different color-bit depths, ranging from 8 (original) to 1.

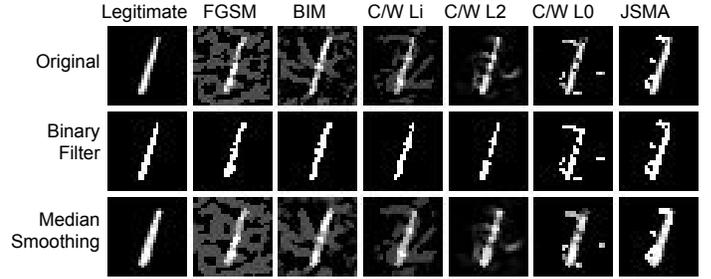

Fig. 3: Examples of adversarial attacks and feature squeezing methods extracted from the MNIST dataset. The first column shows the original image and its squeezed versions, while the other columns present the adversarial variants. All targeted attacks are targeted-next.

- $L_2$: $\|\mathbf{z}\|_2 = \sqrt{\sum_i z_i^2}$.

  The $L_2$ norm corresponds to the Euclidean distance between $\mathbf{x}$ and $\mathbf{x}'$. This distance can remain small when many small changes are applied to many pixels.

- $L_0$: $\|\mathbf{z}\|_0 = \#\{i \mid z_i \neq 0\}$.

  For images, this metric the total number of pixels that may be altered between $\mathbf{x}$ and $\mathbf{x}'$, but not the amount of perturbation.

Here $\mathbf{z} = \mathbf{x} - \mathbf{x}'$. Next, we discuss the eleven attacking algorithms used in our experiments, grouped by the norm they used for $\Delta(\cdot)$.

*1) Fast Gradient Sign Method: FGSM ($L_\infty$, Untargeted)*

Goodfellow et al. hypothesized that DNNs are vulnerable to adversarial perturbations because of their linear nature [10]. They proposed the *fast gradient sign method* (FGSM) for efficiently finding adversarial examples. To control the cost of attacking, FGSM assumes that the attack strength at every feature dimension is the same, essentially measuring the perturbation $\Delta(\mathbf{x}, \mathbf{x}')$ using the $L_\infty$-norm. The strength of perturbation at every dimension is limited by the same constant parameter, $\epsilon$, which is also used as the amount of perturbation.

As an untargeted attack, the perturbation is calculated directly by using gradient vector of a loss function:

$$\Delta(\mathbf{x}, \mathbf{x}') = \epsilon \cdot \text{sign}(\nabla_\mathbf{x} J(g(\mathbf{x}), y)) \quad (3)$$

Here the loss function, $J(\cdot, \cdot)$, is the loss that have been used for training the specific DNN model, and $y$ is the correct label for $\mathbf{x}$. Equation (3) essentially increases the loss $J(\cdot, \cdot)$ by perturbing the input $\mathbf{x}$ based on a transformed gradient.

*2) Basic Iterative Method: BIM ($L_\infty$, Untargeted)*

Kurakin et al. extended the FGSM method by applying it multiple times with small step size [19]. This method clips pixel values of intermediate results after each step to ensure that they are in an $\epsilon$-neighborhood of the original image $\mathbf{x}$. For the $m^\text{th}$ iteration,

$$\mathbf{x}'_{m+1} = \mathbf{x}'_m + \text{Clip}_{\mathbf{x},\epsilon}(\alpha \cdot \text{sign}(\nabla_\mathbf{x} J(g(\mathbf{x}'_m), y))) \quad (4)$$

The clipping equation, $\text{Clip}_{\mathbf{x},\epsilon}(\mathbf{z})$, performs per-pixel clipping on $\mathbf{z}$ so the result will be in the $L_\infty$ $\epsilon$-neighborhood of $\mathbf{x}$ [19].

*3) DeepFool ($L_2$, Untargeted)*

Moosavi et al. used a $L_2$ minimization-based formulation, termed DeepFool, to search for adversarial examples [28]:

$$\Delta(\mathbf{x}, \mathbf{x}') := arg\,min_\mathbf{z}\|\mathbf{z}\|_2, \text{ subject to: } g(\mathbf{x} + \mathbf{z}) \neq g(\mathbf{x}) \quad (5)$$

DeepFool searches for the minimal perturbation to fool a classifier and uses concepts from geometry to direct the search. For linear classifiers (whose decision boundaries are linear planes), the region of the space describing a classifier's output can be represented by a polyhedron (whose plane faces are those boundary planes defined by the classifier). Then DeepFool searches within this polyhedron for the minimal perturbation that can change the classifiers decision. For general non-linear classifiers, this algorithm uses an iterative linearization procedure to get an approximated polyhedron.

*4) Jacobian Saliency Map Approach: JSMA ($L_0$, Targeted)*

Papernot et al. [34] proposed the *Jacobian-based saliency map approach* (JSMA) to search for adversarial examples by only modifying a limited number of input pixels in an image. As a targeted attack, JSMA iteratively perturbs pixels in an input image that have high adversarial saliency scores. The adversarial saliency map is calculated from the Jacobian (gradient) matrix $\nabla_\mathbf{x} g(\mathbf{x})$ of the DNN model $g(\mathbf{x})$ at the current input $\mathbf{x}$. The $(c, p)^\text{th}$ component in Jacobian matrix $\nabla_\mathbf{x} g(\mathbf{x})$ describes the derivative of output class $c$ with respect to feature pixel $p$. The adversarial saliency score of each pixel is calculated to reflect how this pixel will increase the output score of the target class $t$ versus changing the score of all other possible output classes. The process is repeated until classification into the target class is achieved, or it reaches the maximum number of perturbed pixels. Essentially, JSMA optimizes Equation (2) by measuring perturbation $\Delta(\mathbf{x}, \mathbf{x}')$ through the $L_0$-norm.

*5) Carlini/Wagner Attacks ($L_2$, $L_\infty$ and $L_0$, Targeted)*

Carlini and Wagner recently introduced three new gradient-based attack algorithms that are more effective than all previously-known methods in terms of the adversarial success rates achieved with minimal perturbation amounts [7]. They proposed three versions of attacks using $L_2$, $L_\infty$, and $L_0$ norms.

The $CW_2$ attack formalizes the task of generating adversarial examples as an optimization problem with two terms as usual: the prediction objective and the distance term. However, it makes the optimization easier to solve through several techniques. The first is using the logits-based objective function instead of the softmax-cross-entropy loss that is commonly used in other optimization-based attacks. This makes it robust against the defensive distillation method [36]. The second is converting the target variable to the *argtanh* space to bypass the box-constraint on the input, making it more flexible in taking advantage of modern optimization solvers, such as Adam. It also uses a binary search algorithm to select a suitable coefficient that performs a good trade-off between the prediction and the distance terms. These improvements enable the $CW_2$ attack to find adversarial examples with smaller perturbations than previous attacks.

The $CW_\infty$ attack recognizes the fact that $L_\infty$ norm is hard to optimize and only the maximum term is penalized. Thus, it revises the objective into limiting perturbations to be less than a threshold $\tau$ (initially 1, decreasing in each iteration). The optimization reduces $\tau$ iteratively until no solution can be found. Consequently, the resulting solution has all the perturbations smaller than the specified $\tau$.

The basic idea of the $CW_0$ attack is to iteratively use $CW_2$ to find the least important features and freeze them (so value will never be changed) until the $L_2$ attack fails with too many features being frozen. As a result, only those features with significant impact on the prediction are changed. This is the opposite of JSMA, which iteratively selects the most important features and performs large perturbations until it successfully fools the target classifier.

*C. Defensive Techniques*

Papernot et al. [35] provided a comprehensive summary of work on defending against adversarial samples, grouping work into two broad categories: *adversarial training* and *gradient masking*, which we discuss further below. A third approach is to transform the input so that the model is not sensitive to small perturbations. Our proposed feature squeezing method is broadly part of this theme.

**Adversarial Training.** *Adversarial training* introduces discovered adversarial examples and the corresponding ground truth labels to the training [10, 40, 21]. Ideally, the model will learn how to restore the ground truth from the adversarial perturbations and perform robustly on the future adversarial examples. This technique, however, suffers from the high cost to generate adversarial examples and (at least) doubles the training cost of DNN models due to its iterative re-training procedure. Its effectiveness also depends on having a technique for efficiently generating adversarial examples similar to the one used by the adversary, which may not be the case in practice. As pointed out by Papernot et al. [35], it is essential to include adversarial examples produced by all known attacks in adversarial training, since this defensive training is non-adaptive. But, it is computationally expensive to find adversarial inputs by most known techniques, and there is no way to be confident the adversary is limited to techniques that are known to the trainer.

Madry et al. proposed a variation of adversarial training by enlarging the model capacity in the re-training. Their adversarial training uses the adversarial examples generated by BIM attack with random starts, named as the "PGD attack" [21]. The authors claimed that this method could provide a security guarantee against any adversary based on the theory of robust optimization. The empirical results showed that the PGD-based adversarial training significantly increases the robustness of an MNIST model against many different attacks. However, the result on the CIFAR-10 dataset is inferior. We compare our feature squeezing technique with this adversarial training in Section 4-B.

**Gradient Masking.** By forcing DNN models to produce near-zero gradients, the "gradient masking" defenses seek to reduce the sensitivity of DNN models to small changes in inputs. Gu et al. proposed adding a gradient penalty term in the training objective. The penalty term is a summation of the layer-by-layer Frobenius norm of the Jacobian matrix [12]. Although the trained model behaves more robust against adversaries, the penalty significantly reduces the capacity of the model and sacrifices accuracy on many tasks [35]. Papernot et al. introduced the strategy of "defensive distillation" to harden DNN models [36]. A defensively distilled model is trained with the smoothed labels produced by an existing trained DNN model. Then, to hide model's gradient information from an adversary, the distilled model replaces its last layer with a "harder" softmax function after training. Experimental results showed that larger perturbations are required when using JSMA to evade distilled models. However, two subsequent studies have found that defensive distillation failed to mitigate a variant of JSMA with a division trick [6] and a black-box attack [33]. Papernot et al. concluded that methods designed to conceal gradient information are bound to have limited success because of the transferability of adversarial examples [35].

**Input Transformation.** A few recent studies for hardening deep learning try to reduce the model sensitivity to small input changes by transforming the inputs. Bhagoji et al. proposed to use dimensionality reduction techniques such as Principal Component Analysis (PCA) as defense [2]. They first performed PCA on a clean dataset, then linearly projected all the inputs to the PCA space and only preserved the top *k* principle axes. While we could expect the reduced sensitivity with the PCA projection, the method corrupts the spatial structure of an image, and the state-of-the-art CNN models are no longer applicable. Instead, Meng and Chen proposed to train an autoencoder as an image filter to harden DNN models [24]. The encoder stage of the autoencoder is essentially a non-linear dimensionality reduction. Its decoder stage restores an input to its original spatial structure; therefore the target DNN model does not need to change. Similar to ours, Osadchy et al. independently suggested using the binary filter and the median smoothing filter to eliminate adversarial perturbations and proposed to attack the defenses by increasing attackers' perturbation strength [31]. Differently, we focus on the task of detecting adversarial examples and show that increasing the perturbation amount may result in unrecognizable images in Section 5-D.

*D. Detecting Adversarial Examples*

Multiple recent studies [25, 11, 9] focused on detecting adversarial examples. The strategies they explored naturally fall into three groups: *sample statistics*, *training a detector* and *prediction inconsistency*. Our proposed feature squeezing belongs to the third group that employs prediction inconsistency.

**Sample Statistics.** For detecting adversarial examples, Grosse et al. [11] proposed a statistical test using maximum mean discrepancy and suggests the energy distance as the statistical distance measure. Their method requires a large set of both adversarial and legitimate inputs and is not capable of detecting individual adversarial examples, making it not useful in practice. Feinman et al. proposed to use kernel density estimation [9] that measures the distance between an unknown input and a group of legitimate inputs using their representations from some middle layers of a DNN model. It is computationally expensive and can only detect adversarial examples lying far from the manifolds of the legitimate population. Due to the intrinsically unperceptive nature of adversarial examples, using sample statistics to separate adversarial examples from legitimate inputs seems unlikely to be effective. Experimental results from both Grosse et al. [11] and Feinman et al. [9] showed that strategies relying on sample statistics gave inferior detection performance compared to other detection methods.

**Training a Detector.** Similar to adversarial training, adversarial examples can be used to train a detector. However, this strategy requires a large number of adversarial examples, therefore, being expensive and prone to overfitting the adversarial attacks that generated examples for training the detector. Metzen et al. proposed attaching a CNN-based detector as a branch off a middle layer of the original DNN model [25]. The detector outputs two classes and uses adversarial examples (as one class) plus legitimate examples (as the other class) for training. The detector is trained while freezing the weights of the original DNN, therefore does not sacrifice the classification accuracy on the legitimate inputs. Grosse et al. demonstrated a detection method (previously proposed by Nguyen et al. [29]) that adds a new "adversarial" class in the last layer of the DNN model [11]. The revised model is trained with both legitimate and adversarial inputs, reducing the accuracy on legitimate inputs due to the change to the model architecture.

**Prediction Inconsistency.** The basic idea of *prediction inconsistency* is to measure the disagreement among several models in predicting an unknown input, since one adversarial example may not fool every DNN model. Feinman et al. borrowed an idea from dropout [14] and designed a detection technique they called *Bayesian neural network uncertainty* [9]. The authors used the "training" mode of dropout layers to generate many predictions for an input at test time. They reported that the disagreement among the predictions of sub-models is rare on legitimate inputs but common on adversarial examples, thus can be used for detection. Meng and Chen independently proposed a similar adversary detection method to ours that also uses the prediction vectors of the original and the filtered images [24]. The biggest difference is that they trained an autoencoder as the image filter, whereas we rely on "hard-coded" transformations. As a result, our approach is less expensive in the training phase. We compare the performance of their detectors with ours in Table 5.

## 3. Feature Squeezing Methods

Although the notion of feature squeezing is quite general, we focus on two simple types of squeezing: reducing the color depth of images (Section 3-A), and using smoothing (both local and non-local) to reduce the variation among pixels (Section 3-B). Section 4 looks at the impact of each squeezing method on classifier accuracy and robustness against adversarial inputs. These results enable feature squeezing to be used for detecting adversarial examples in Section 5.

### A. Color Depth

A neural network, as a differentiable model, assumes that the input space is continuous. However, digital computers only support discrete representations as approximations of continuous natural data. A standard digital image is represented by an array of pixels, each of which is usually represented as a number that represents a specific color.

Common image representations use color bit depths that lead to irrelevant features, so we hypothesize that reducing bit depth can reduce adversarial opportunity without harming classifier accuracy. Two common representations, which we focus on here because of their use in our test datasets, are 8-bit grayscale and 24-bit color. A grayscale image provides $2^8 = 256$ possible values for each pixel. An 8-bit value represents the intensity of a pixel where 0 is black, 255 is white, and intermediate numbers represent different shades of gray. The 8-bit scale can be extended to display color images with separate red, green and blue color channels. This provides 24 bits for each pixel, representing $2^{24} \approx 16$ million different colors.

#### 1) Squeezing Color Bits

While people usually prefer larger bit depth as it makes the displayed image closer to the natural image, large color depths are often not necessary for interpreting images (for example, people have no problem recognizing most black-and-white images). We investigate the bit depth squeezing with three popular datasets for image classification: MNIST, CIFAR-10 and ImageNet.

**Greyscale Images (MNIST).** The MNIST dataset contains 70,000 images of hand-written digits (0 to 9). Of these, 60,000 images are used as training data and the remaining 10,000 images are used for testing. Each image is $28 \times 28$ pixels, and each pixel is encoded as 8-bit grayscale.

Figure 2 shows one example of class 0 in the MNIST dataset in the first row, with the original 8-bit grayscale images in the leftmost and the 1-bit monochrome images rightmost. The rightmost images, generated by applying a binary filter with 0.5 as the cutoff, appear nearly identical to the original images on the far left. The processed images are still recognizable to humans, even though the feature space is only $1/128^{th}$ the size of the original 8-bit grayscale space.

Figure 3 hints at why reducing color depth can mitigate adversarial examples generated by multiple attack techniques. The top row shows one original example of class 1 from the MNIST test set and six different adversarial examples. The middle row shows those examples after reducing the bit depth of each pixel into binary. To a human eye, the binary-filtered images look more like the correct class; in our experiments, we find this is true for DNN classifiers also (Table 3 in Section 4).

**Color Images (CIFAR-10 and ImageNet).** We use two datasets of color images in this paper: the CIFAR-10 dataset with tiny images and the ImageNet dataset with high-resolution photographs. The CIFAR-10 dataset contains 60,000 images, each with 32 × 32 pixels encoded with 24-bit color and belonging to 10 different classes. The ImageNet dataset is provided by ImageNet Large Scale Visual Recognition Challenge 2012 for the classification task, which contains 1.2 million training images and the other 50,000 images for validation. The photographs in the ImageNet dataset are in different sizes and hand-labeled with 1,000 classes. However, they are pre-processed to 224×224 pixels encoded with 24-bit True Color for the target model MobileNet [16, 23] we use in this paper.

The middle row and the bottom row of Figure 2 show that we can reduce the original 8-bit (per RGB channel) images to fewer bits without significantly decreasing the image recognizability to humans. It is difficult to tell the difference between the original images with 8-bit per channel color and images using as few as 4 bits of color depth. Unlike what we observed in the MNIST datase, however, bit depths lower than 4 do introduce some human-observable loss. This is because we lose much more information in the color image even though we reduce to the same number of bits per channel. For example, if we reduce the bits-per-channel from 8 bits to 1 bit, the resulting grayscale space is 1/128 large as the original; the resulting RGB space is only $2^{-(24-3)} = 1/2,097,152$ of the original size. Nevertheless, in Section 4-A we find that squeezing to 4 bits is strong enough to mitigate a lot of adversarial examples while preserving the accuracy on legitimate examples.

*2) Implementation*

We implement the bit depth reduction operation in Python with the NumPy library. The input and output are in the same numerical scale [0, 1] so that we don't need to change anything of the target models. For reducing to $i$-bit depth ($1 \leq i \leq 7$), we first multiply the input value with $2^i - 1$ (minus 1 due to the zero value) then round to integers. Next we scale the integers back to [0, 1], divided by $2^i - 1$. The information capacity of the representation is reduced from 8-bit to $i$-bit with the integer-rounding operation.

### B. Spatial Smoothing

Spatial smoothing (also known as *blur*) is a group of techniques widely used in image processing for reducing image noise. Next, we describe the two types of spatial smoothing methods we used: *local smoothing* and *non-local smoothing*.

*1) Local Smoothing*

Local smoothing methods make use of the nearby pixels to smooth each pixel. By selecting different mechanisms in weighting the neighbouring pixels, a local smoothing method can be designed as Gaussian smoothing, mean smoothing or the median smoothing method [38] we use. As we report in Section 4-A, median smoothing (also known as *median blur* or *median filter*) is particularly effective in mitigating adversarial examples generated by $L_0$ attacks.

The median filter runs a sliding window over each pixel of the image, where the center pixel is replaced by the median value of the neighboring pixels within the window. It does not actually reduce the number of pixels in the image, but spreads pixel values across nearby pixels. The median filter is essentially squeezing features out of the sample by making adjacent pixels more similar.

The size of the window is a configurable parameter, ranging from 1 up to the image size. If it were set to the image size, it would (modulo edge effects) flatten the entire image to one color. A square shape window is often used in median filtering, though there are other design choices. Several padding methods can be employed for the pixels on the edge, since there are no real pixels to fill the window. We choose *reflect padding*, in which we mirror the image along with the edge for calculating the median value of a window when necessary.

Median smoothing is particularly effective at removing sparsely-occurring black and white pixels in an image (descriptively known as *salt-and-pepper noise*), whilst preserving edges of objects well.

**Implementation.** We use the median filter implemented in SciPy [38]. In a 2×2 sliding window, the center pixel is always located in the lower right. When there are two equal-median values due to the even number of pixels in a window, we (arbitrarily) use the greater value as the median.

*2) Non-local Smoothing*

Non-local smoothing is different from local smoothing because it smooths over similar pixels in a much larger area instead of just nearby pixels. For a given image patch, non-local smoothing finds several similar patches in a large area of the image and replaces the center patch with the average of those similar patches. Assuming that the mean of the noise is zero, averaging the similar patches will cancel out the noise while preserving the edges of an object. Similar with local smoothing, there are several possible ways to weigh the similar patches in the averaging operation, such as Gaussian, mean, and median. We use a variant of the Gaussian kernel because it is widely used and allows to control the deviation from the mean. The parameters of a non-local smoothing method typically include the search window size (a large area for searching similar patches), the patch size and the filter strength (bandwidth of the Gaussian kernel). We will denote a filter as "non-local means (a-b-c)" where "a" means the search window $a \times a$, "b" means the patch size $b \times b$ and "c" means the filter strength.

**Implementation.** We use the fast non-local means denoising method implemented in OpenCV [30]. It first converts a color image to the CIELAB colorspace, then separately denoises its L and AB components, then converts back to the RGB space.

### C. Other Squeezing Methods

Our results in this paper are limited to these simple squeezing methods, which are surprisingly effective on our test datasets. However, we believe many other squeezing methods are possible, and continued experimentation will be worthwhile to find the most effective squeezing methods.

One possible area to explore includes lossy compression techniques. Kurakin et al. explored the effectiveness of the JPEG format in mitigating the adversarial examples [19]. Their experiment shows that a very low JPEG quality (e.g. 10 out of 100) is able to destruct the adversarial perturbations generated by FGSM with $\epsilon$=16 (at scale of [0,255]) for at most 30% of the successful adversarial examples. However, they didn't evaluate the potential loss on the accuracy of legitimate inputs.

Another possible direction is dimension reduction. For example, Turk and Pentland's early work pointed out that many pixels are irrelevant features in the face recognition tasks, and the face images can be projected to a feature space named *eigenfaces* [41]. Even though image samples represented in the *eigenface*-space loose the spatial information a CNN model needs, the image restoration through *eigenfaces* may be a useful technique to mitigate adversarial perturbations in a face recognition task.

## 4. Robustness

The previous section suggested that images, as used in classification tasks, contain many irrelevant features that can be squeezed without reducing recognizability. For feature squeezing to be effective in detecting adversarial examples (Figure 1), it must satisfy two properties: (1) on adversarial examples, the squeezing reverses the effects of the adversarial perturbations; and (2) on legitimate examples, the squeezing does not significantly impact a classifier's predictions. This section evaluates the how well different feature squeezing methods achieve these properties.

**Threat model.** In evaluating robustness, we assume a powerful adversary who has full access to a trained target model, but no ability to influence that model. For now, we assume the adversary is not aware of feature squeezing being performed on the operator's side. The adversary tries to find inputs that are misclassified by the model using white-box attack techniques.

Although we analyze the robustness of standalone feature squeezers here, we do not propose using a standalone squeezer as a defense because an adversary may take advantage of feature squeezing in attacking a DNN model. For example, when facing binary squeezing, an adversary can construct an image by setting all pixel intensity values to be near 0.5. This image is entirely gray to human eyes. By setting pixel values to either 0.49 or 0.51 it can result in an arbitrary 1-bit filtered image after squeezing, either entirely white or black. Such an attack can easily be detected by our detection framework (Section 5), however. Section 5-D considers how adversaries can adapt to our detection framework.

**Target Models.** We use three popular datasets for the image classification task: MNIST, CIFAR-10, and ImageNet. For each dataset, we set up a pre-trained model with the state-of-the-art performance. Table 1 summarizes the prediction performance of each model and its DNN architecture. Our MNIST model (a seven-layer CNN [4]) and CIFAR-10 model (a DenseNet [17, 22] model) both achieve prediction performance competitive with state-of-the-art results [1]. For the ImageNet dataset, we use a MobileNet model [16, 23] because MobileNets are more widely used on mobile phones and their small and efficient design make it easier to conduct experiments. The accuracy

TABLE 1: Summary of the target DNN models.

| Dataset | Model | Top-1 Accuracy | Top-1 Mean Confidence | Top-5 Accuracy |
|---|---|---|---|---|
| MNIST | 7-Layer CNN [4] | 99.43% | 99.39% | - |
| CIFAR-10 | DenseNet [17, 22] | 94.84% | 92.15% | - |
| ImageNet | MobileNet [16, 23] | 68.36% | 75.48% | 88.25% |

TABLE 2: Evaluation of attacks.

| | Configuration | | | Cost (s) | Success Rate | Prediction Confidence | Distortion | | |
|---|---|---|---|---|---|---|---|---|---|
| | Attack | Mode | | | | | $L_\infty$ | $L_2$ | $L_0$ |
| MNIST | $L_\infty$ | FGSM | | 0.002 | 46% | 93.89% | 0.302 | 5.905 | 0.560 |
| | | BIM | | 0.01 | 91% | 99.62% | 0.302 | 4.758 | 0.513 |
| | | $CW_\infty$ | Next | 51.2 | 100% | 99.99% | 0.251 | 4.091 | 0.491 |
| | | | LL | 50.0 | 100% | 99.98% | 0.278 | 4.620 | 0.506 |
| | $L_2$ | $CW_2$ | Next | 0.3 | 99% | 99.23% | 0.656 | 2.866 | 0.440 |
| | | | LL | 0.4 | 100% | 99.99% | 0.734 | 3.218 | 0.436 |
| | $L_0$ | $CW_0$ | Next | 68.8 | 100% | 99.99% | 0.996 | 4.538 | 0.047 |
| | | | LL | 74.5 | 100% | 99.99% | 0.996 | 5.106 | 0.060 |
| | | JSMA | Next | 0.8 | 71% | 74.52% | 1.000 | 4.328 | 0.047 |
| | | | LL | 1.0 | 48% | 74.80% | 1.000 | 4.565 | 0.053 |
| CIFAR-10 | $L_\infty$ | FGSM | | 0.02 | 85% | 84.85% | 0.016 | 0.863 | 0.997 |
| | | BIM | | 0.2 | 92% | 95.29% | 0.008 | 0.368 | 0.993 |
| | | $CW_\infty$ | Next | 225 | 100% | 98.22% | 0.012 | 0.446 | 0.990 |
| | | | LL | 225 | 100% | 97.79% | 0.014 | 0.527 | 0.995 |
| | $L_2$ | DeepFool | | 0.4 | 98% | 73.45% | 0.028 | 0.235 | 0.995 |
| | | $CW_2$ | Next | 10.4 | 100% | 97.90% | 0.034 | 0.288 | 0.768 |
| | | | LL | 12.0 | 100% | 97.35% | 0.042 | 0.358 | 0.855 |
| | $L_0$ | $CW_0$ | Next | 367 | 100% | 98.19% | 0.650 | 2.103 | 0.019 |
| | | | LL | 426 | 100% | 97.60% | 0.712 | 2.530 | 0.024 |
| | | JSMA | Next | 8.4 | 100% | 43.29% | 0.896 | 4.954 | 0.079 |
| | | | LL | 13.6 | 98% | 39.75% | 0.904 | 5.488 | 0.098 |
| ImageNet | $L_\infty$ | FGSM | | 0.02 | 99% | 63.99% | 0.008 | 3.009 | 0.994 |
| | | BIM | | 0.2 | 100% | 99.71% | 0.004 | 1.406 | 0.984 |
| | | $CW_\infty$ | Next | 211 | 99% | 90.33% | 0.006 | 1.312 | 0.850 |
| | | | LL | 269 | 99% | 81.42% | 0.010 | 1.909 | 0.952 |
| | $L_2$ | DeepFool | | 60.2 | 89% | 79.59% | 0.027 | 0.726 | 0.984 |
| | | $CW_2$ | Next | 20.6 | 90% | 76.25% | 0.019 | 0.666 | 0.323 |
| | | | LL | 29.1 | 97% | 76.03% | 0.031 | 1.027 | 0.543 |
| | $L_0$ | $CW_0$ | Next | 608 | 100% | 91.78% | 0.898 | 6.825 | 0.003 |
| | | | LL | 979 | 100% | 80.67% | 0.920 | 9.082 | 0.005 |

Results are for 100 seed images for each the DNN models described in Table 1. The *cost* of an attack generating adversarial examples is measured in seconds per sample. The $L_0$ distortion is normalized by the number of pixels (e.g., 0.560 means 56% of all pixels in the image are modified).

achieved by this model (88.25% top-5 accuracy), is below what can be achieved with a larger model such as Inception v3 [39, 8] (93.03% top-5 accuracy).

**Attacks.** We evaluate feature squeezing on all of the attacks described in Section 2-C and summarized in Table 2. For each targeted attack, we try two different targets: the *Next* class ($t = L + 1$ mod #*classes*), and the least-likely class (*LL*), $t = \min(\hat{y})$. Here $t$ is the target class, $L$ is the index of the ground-truth class and $\hat{y}$ is the prediction vector of an input image. This gives eleven total attacks: the three untargeted attacks (FGSM, BIM and DeepFool), and two versions each of the four targeted attacks (JSMA, $CW_\infty$, $CW_2$, and $CW_0$). We use the implementations of FGSM, BIM and JSMA provided by the Cleverhans library [32]. For DeepFool and the three CW attacks, we use the implementations from the original authors [4, 27]. Our models and code for our attacks, defenses, and tests are available at https://evadeML.org/zoo. We use a PC equipped with an i7-6850K 3.60GHz CPU, 64GiB system memory, and a GeForce GTX 1080 to conduct the experiments.

For the seed images, we select the first 100 correctly predicted examples in the test (or validation) set from each dataset for all the attack methods, since some attacks are too expensive

TABLE 3: Model accuracy with feature squeezing

| Dataset | Squeezer | | $L_\infty$ Attacks | | | | $L_2$ Attacks | | | $L_0$ Attacks | | | | All Attacks | Legitimate |
|---|---|---|---|---|---|---|---|---|---|---|---|---|---|---|---|
| | Name | Parameters | FGSM | BIM | CW$_\infty$ | | Deep-Fool | CW$_2$ | | CW$_0$ | | JSMA | | | |
| | | | | | Next | LL | | Next | LL | Next | LL | Next | LL | | |
| MNIST | None | | 54% | 9% | 0% | 0% | - | 0% | 0% | 0% | 0% | 27% | 40% | 13.00% | 99.43% |
| | Bit Depth | 1-bit | 92% | 87% | 100% | 100% | - | 83% | 66% | 0% | 0% | 50% | 49% | 62.70% | 99.33% |
| | Median Smoothing | 2x2 | 61% | 16% | 70% | 55% | - | 51% | 35% | 39% | 36% | 62% | 56% | 48.10% | 99.28% |
| | | 3x3 | 59% | 14% | 43% | 46% | - | 51% | 53% | 67% | 59% | 82% | 79% | 55.30% | 98.95% |
| CIFAR-10 | None | | 15% | 8% | 0% | 0% | 2% | 0% | 0% | 0% | 0% | 0% | 0% | 2.27% | 94.84% |
| | Bit Depth | 5-bit | 17% | 13% | 12% | 19% | 40% | 40% | 47% | 0% | 0% | 21% | 17% | 20.55% | 94.55% |
| | | 4-bit | 21% | 29% | 69% | 74% | 72% | 84% | 84% | 7% | 10% | 23% | 20% | 44.82% | 93.11% |
| | Median Smoothing | 2x2 | 38% | 56% | 84% | 86% | 83% | 87% | 83% | 88% | 85% | 84% | 76% | 77.27% | 89.29% |
| | Non-local Means | 11-3-4 | 27% | 46% | 80% | 84% | 76% | 84% | 88% | 11% | 11% | 44% | 32% | 53.00% | 91.18% |
| ImageNet | None | | 1% | 0% | 0% | 0% | 11% | 10% | 3% | 0% | 0% | - | - | 2.78% | 69.70% |
| | Bit Depth | 4-bit | 5% | 4% | 66% | 79% | 44% | 84% | 82% | 38% | 67% | - | - | 52.11% | 68.00% |
| | | 5-bit | 2% | 0% | 33% | 60% | 21% | 68% | 66% | 7% | 18% | - | - | 30.56% | 69.40% |
| | Median Smoothing | 2x2 | 22% | 28% | 75% | 81% | 72% | 81% | 84% | 85% | 85% | - | - | 68.11% | 65.40% |
| | | 3x3 | 33% | 41% | 73% | 76% | 66% | 77% | 79% | 81% | 79% | - | - | 67.22% | 62.10% |
| | Non-local Means | 11-3-4 | 10% | 25% | 77% | 82% | 57% | 87% | 86% | 43% | 47% | - | - | 57.11% | 65.40% |

No results are shown for DeepFool on MNIST because of the adversarial examples it generates appear unrecognizable to humans; no results are shown for JSMA on ImageNet because it requires more memory than available to run. We do not apply the non-local smoothing on MNIST images because it is difficult to find similar patches on such images for smoothing a center patch.

to run on all the seeds. We adjust the applicable parameters of each attack to generate high-confidence adversarial examples, otherwise they would be easily rejected. This is because the three DNN models we use achieve high confidence of the top-1 predictions on legitimate examples (see Table 1; mean confidence is over 99% for MNIST, 92% for CIFAR-10, and 75% for ImageNet). In addition, all the pixel values in the generated adversarial images are clipped and squeezed to 8-bit-per-channel pixels so that the resulting inputs are within the possible space of images.

Table 2 reports results from our evaluation of the eleven attacks on three datasets. The success rate captures the probability an adversary achieves their goal. For untargeted attacks, the reported success rate is $1-accuracy$; for targeted attacks, it is only considered a success if the model predicts the targeted class. Most of the attacks are very effective in generating high-confidence adversarial examples against three DNN models. The CW attacks often produce smaller distortions than other attacks using the same norm objective, but are much more expensive to generate. On the other hand, FGSM, DeepFool, and JSMA often produce low-confidence adversarial examples. We exclude the DeepFool attack from the MNIST dataset because it generates images that appear unrecognizable to humans.[1] We do not have JSMA results for ImageNet because our 64GiB test machine runs out of memory.

### A. Results

Table 3 summarizes the effectiveness of different feature squeezers on classification accuracy in our experiments.

**Color Depth Reduction.** The resolution of a specific bit depth is defined as the number of possible values for each pixel. For example, the resolution of 8-bit color depth is 256. Reducing the bit depth lowers the resolution and diminishes the opportunity an adversary has to find effective perturbations.

*MNIST.* The last column of Table 3 shows the binary filter (1-bit depth reduction) barely reduces the accuracy on the legitimate examples of MNIST (from 99.43% to 99.33% on the test set). The binary filter is effective on all the $L_2$ and $L_\infty$ attacks. For example, it improves the accuracy on CW$_\infty$ adversarial examples from 0% to 100%. The binary filter works well even for large $L_\infty$ distortions. This is because the binary filter squeezes each pixel into 0 or 1 using a cutoff 0.5 in the $[0, 1)$ scale. This means maliciously perturbing a pixel's value by ±0.30 does not change the squeezed value of pixels whose original values fall into $[0, .20)$ and $[.80, 1)$. In contrast, bit depth reduction is not effective against $L_0$ attacks (JSMA and CW$_0$) since these attacks make large changes to a few pixels that are not reversed by the bit depth squeezer. However, as we will show later, the spatial smoothing squeezers are often effective against $L_0$ attacks.

*CIFAR-10 and ImageNet.* Because the DNN models for CIFAR-10 and ImageNet are more sensitive to perturbations, adversarial examples at very low $L_2$ and $L_\infty$ distortions can be found. Table 3 includes the results of 4-bit depth and 5-bit depth filters in mitigating the adversaries for CIFAR-10 and ImageNet. In testing, the 5-bit depth filter increases the accuracy on adversarial inputs for several of the attacks (for example, increasing accuracy from 0% to 40% for the CW$_2$ Next class targeted attack), while almost perfectly preserving the accuracy on legitimate data (94.55% compared with 94.84%). The more aggressive 4-bit depth filter is more robust against adversaries (e.g., accuracy on CW$_2$ increases to 84%), but reduces the accuracy on legitimate inputs from 94.84% to 93.11%. We do not believe these results are good enough for use as a stand-alone defense (even ignoring the risk of adversarial adaptation), but they provide some insight why the method is effective as used in our detection framework.

**Median Smoothing.** The adversarial perturbations produced by the $L_0$ attacks (JSMA and CW$_0$) are similar to *salt-and-pepper noise*, though it is introduced intentionally instead of randomly. Note that the adversarial strength of an $L_0$ adversary limits the number of pixels that can be manipulated, so it is not surprising that maximizing the amount of change to each modified pixel is typically most useful to the adversary. This is why the smoothing squeezers are more effective against these attacks than the color depth squeezers.

---
[1] We believe this is because the linear boundary assumption doesn't hold for the particular MNIST model and DeepFool fails to approximate the minimal perturbation.

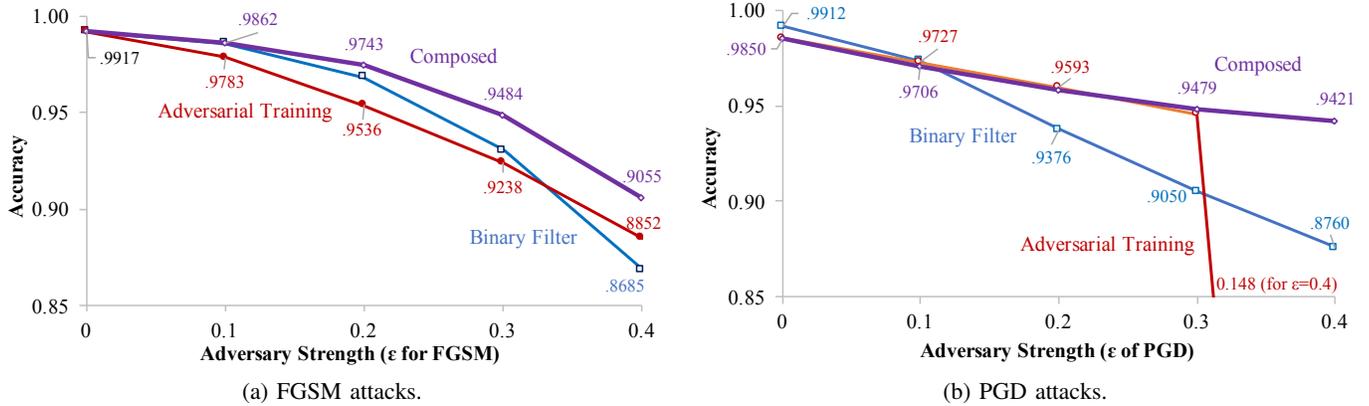

(a) FGSM attacks.

(b) PGD attacks.

Fig. 4: **Composing adversarial training with feature squeezing.** The horizontal axis is the adversary's strength ($\epsilon$), increasing to the right. The adversarial training uses $\epsilon = 0.3$ for both FGSM are PGD. Composing the 1-bit filter with the adversarial-trained model often performs the best.

*MNIST.* We evaluate two window sizes on the MNIST dataset in Table 3. Median smoothing is the best squeezer for all of the $L_0$ attacks (CW$_0$ and JSMA). The median filter with $2 \times 2$ window size performs slightly worse on adversarial examples than the one with $3 \times 3$ window, but it almost perfectly preserves the performance on the legitimate examples (decreasing accuracy from 99.43% to 99.28%).

*CIFAR-10 and ImageNet.* The experiment confirms the intuition that median smoothing can effectively eliminate the $L_0$-limited perturbations. Without squeezing, the $L_0$ attacks are effective on CIFAR-10, resulting in 0% accuracy for the original model ("None" row in Table 3). However, with a $2\times 2$ median filter, the accuracy increases to over 75% for all the four $L_0$ type attacks. We observe similar results on ImageNet, where the accuracy increases from 0% to 85% for the CW$_0$ attacks after median smoothing.

**Non-local Smoothing.** Non-local smoothing has comparable performance in increasing the accuracy on adversarial examples other than the $L_0$ type. On the other hand, it has little impact on the accuracy on legitimate examples. For example, the $2\times 2$ median filter decreases the accuracy on the CIFAR-10 model from 94.84% to 89.29% while the model with non-local smoothing still achieves 91.18%.

*B. Combining with Adversarial Training*

Since our approach modifies inputs rather than the model, it can easily be composed with any defense technique that operates on the model. The most successful previous defense against adversarial examples is adversarial training (Section 2-C). To evaluate the effectiveness of composing feature squeezing with adversarial training, we combined it with two different adversarial training methods: the FGSM-based version implemented in Cleverhans [32] and the PGD-based version implemented by Madry et al. [21]. We evaluated the accuracy on all 10,000 MNIST testing images and compared the three different configurations: 1. the base model with a binary filter; 2. the adversarial-trained model; 3. the adversarial-trained model with a binary filter.

Figure 4a shows that bit-depth reduction by itself often outperforms the adversarial training methods, and the composition is nearly always the most effective. For FGSM, binary filter feature squeezing outperforms adversarial training for $\epsilon$ values ranging from 0.1 to 0.3. This is the best case for adversarial training since the adversarially-trained model is learning from the same exact adversarial method (training is done with FGSM examples generated at $\epsilon = 0.3$) as the one used to produce adversarial examples in the test. Nevertheless, feature squeezing outperforms it, even at $\epsilon = 0.3$ (93.03% accuracy on adversarial examples compared to 92.38%).

PGD-based adversarial training [21] does better, outperforming the simple binary filter feature squeezing for $\epsilon = 0.2$ and 0.3 but slightly reducing accuracy on legitimate ($\epsilon = 0$) examples, as shown in Figure 4b. Composing both methods typically leads to the highest accuracy or the one comparable to the best single approach. For example, both methods encounter significant drop on accuracy when $\epsilon = 0.4$: 87.60% with the binary filter and 14.84% with adversarial training. However, the composed method still achieves 94.21%.

Feature squeezing is far less expensive than adversarial training. It is almost cost-free, as we simply insert a binary filter before the pre-trained MNIST model. On the other hand, adversarial training is very expensive as it requires both generating adversarial examples and retraining the classifier for many epochs. When its cost is not prohibitive, though, adversarial training is still beneficial since it can be combined with feature squeezing.

## 5. DETECTING ADVERSARIAL INPUTS

The results from Section 4 show that feature squeezing is capable of obtaining accurate model predictions for many adversarial examples with little reduction in accuracy for legitimate examples. This enables detection of adversarial inputs using the framework introduced in Figure 1. The basic idea is to compare the model's prediction on the original sample with the same model's prediction on the sample after squeezing. The model's predictions for a legitimate example and its squeezed version should be similar. On the contrary, if the original and squeezed examples result in dramatically different predictions, the input is likely to be adversarial. Figure 5 confirms this intuition visually by comparing the $L_1$ distances between the predictions for squeezed and non-squeezed examples for legitimate and adversarial examples, and Table 4 shows the results of our experiments. The following subsections provide more details on our detection method, experimental setup, and discuss the results. Section 5-D considers how adversaries may adapt to our defense.

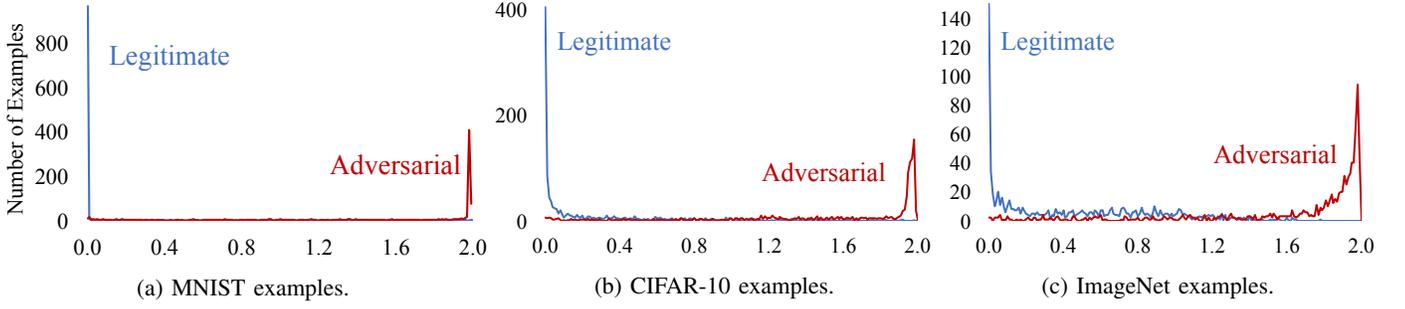

Fig. 5: Differences in $L_1$ distance between original and squeezed sample, for legitimate and adversarial examples across three datasets. The $L_1$ score has a range from 0.0 to 2.0. Each curve is fitted over 200 histogram bins each representing the $L_1$ distance range of 0.01. Each sample is counted in the bin for the maximum $L_1$ distance between the original prediction and the output of the best joint-detection squeezing configuration shown in Table 4. The curves for adversarial examples are for all adversarial examples, including unsuccessful ones (so the separation for successful ones is even larger than shown here).

### A. Detection Method

A prediction vector generated by a DNN classifier normally represents the probability distribution how likely an input sample is to belong to each possible class. Hence, comparing the model's original prediction with the prediction on the squeezed sample involves comparing two probability distribution vectors. There are many possible ways to compare the probability distributions, such as the $L_1$ norm, the $L_2$ norm and K-L divergence [3]. For this work, we select the $L_1$ norm[2] as a natural measure of the difference between the original prediction vector and the squeezed prediction:

$$score^{(\mathbf{x}, \mathbf{x}_{squeezed})} = \|g(\mathbf{x}) - g(\mathbf{x}_{squeezed})\|_1 \quad (6)$$

Here $g(\mathbf{x})$ is the output vector of a DNN model produced by the softmax layer whose $i^{th}$ entry describes the probability how likely input $\mathbf{x}$ is in the $i^{th}$ class. The $L_1$ score has a range from 0 to 2 for the prediction vectors. Higher scores mean there are greater differences between the original prediction and the squeezed prediction. The maximum value of 2 is reached when each prediction vector consists of a 1 and all zeros, but with different classes as the 1. Based on the accuracy results in Section 4, we expect the score to be small for legitimate inputs and large for adversarial examples. The effectiveness of detection depends on selecting a threshold value that accurately distinguishes between legitimate and adversarial inputs.

Even though we can select an effective feature squeezer for a specific type of adversarial method, an operator typically does not know the exact attack method that would be used in practice. Hence, we combine multiple feature squeezers for

[2]This turned out to work well, but it is certainly worth exploring in future work if other metrics can work better.

detection by outputting the maximum distance:

$$score^{joint} = \max\left(score^{(\mathbf{x}, \mathbf{x}_{sq1})}, score^{(\mathbf{x}, \mathbf{x}_{sq2})}, \dots\right) \quad (7)$$

We chose the max operator based on the assumption that different squeezers will be effective for different types of perturbations, and that the effective squeezer can be identified through the highest $L_1$ score. On the other hand, this may increase the false positive rate because the max operator also selects the most destructive squeezer on legitimate inputs. We found that using max is enough to allow a reasonable trade-off in the empirical results, but it may be worthwhile to investigate better ways to combine the squeezers in future work.

Figure 5 shows the histogram of $score^{joint}$ for both legitimate (blue) and adversarial examples (red) on the three datasets. The peak for legitimate examples is always near 0, and the peak for adversarial examples is always near 2. Picking a threshold value between the two peaks is a balance between high detection rates and acceptable false positive rates.

### B. Experimental Setup

We report on experiments using all attacks from Section 4 with the three types of squeezers in different configurations.

**Datasets.** To get a balanced dataset for detection, we select the same number of legitimate examples from the test (or validation) set of each dataset. For each of the attacks in Section 4, we use the 100 adversarial examples generated for each attack in the robustness experiments. This results in 2,000 total examples for MNIST (of which 1,000 are legitimate examples, and 1,000 are adversarial), 2,200 examples for CIFAR-10 and 1,800 examples for ImageNet. We randomly split each detection dataset into two groups: one-half for training the detector and the remainder for validation. Note that some of the adversarial examples are failed adversarial examples which do not confuse the original model, so the number of successful adversarial examples varies slightly across the attacks.

**Squeezers.** We first evaluate how well each squeezing configuration does against adversarial examples generated by each attack method. Then, we consider the realistic scenario where the defender does not know that attack method used by the adversary and needs to select a configuration that works well against a distribution of possible attacks.

TABLE 4: Detection rate for squeezing configurations on successful adversarial examples.

| | Configuration | | | $L_\infty$ Attacks | | | | $L_2$ Attacks | | | $L_0$ Attacks | | | | Overall Detection Rate |
|---|---|---|---|---|---|---|---|---|---|---|---|---|---|---|---|
| | Squeezer | Parameters | Threshold | FGSM | BIM | CW$_\infty$ Next | CW$_\infty$ LL | Deep Fool | CW$_2$ Next | CW$_2$ LL | CW$_0$ Next | CW$_0$ LL | JSMA Next | JSMA LL | |
| MNIST | Bit Depth | 1-bit | 0.0005 | **1.000** | **0.979** | **1.000** | **1.000** | - | **1.000** | **1.000** | 0.556 | 0.563 | **1.000** | **1.000** | 0.903 |
| | | 2-bit | 0.0002 | 0.615 | 0.064 | 0.615 | 0.755 | - | 0.963 | 0.958 | 0.378 | 0.396 | 0.969 | **1.000** | 0.656 |
| | Median Smoothing | 2x2 | 0.0029 | 0.731 | 0.277 | **1.000** | **1.000** | - | 0.944 | **1.000** | 0.822 | 0.938 | 0.938 | **1.000** | 0.868 |
| | | 3x3 | 0.0390 | 0.385 | 0.106 | 0.808 | 0.830 | - | 0.815 | 0.958 | 0.889 | **1.000** | 0.969 | **1.000** | 0.781 |
| | Best Attack-Specific Single Squeezer | | - | **1.000** | **0.979** | **1.000** | **1.000** | - | **1.000** | **1.000** | 0.889 | **1.000** | **1.000** | **1.000** | - |
| | Best Joint Detection (1-bit, 2x2) | | 0.0029 | **1.000** | **0.979** | **1.000** | **1.000** | - | **1.000** | **1.000** | **0.911** | 0.938 | **1.000** | **1.000** | **0.982** |
| CIFAR-10 | Bit Depth | 1-bit | 1.9997 | 0.063 | 0.075 | 0.000 | 0.000 | 0.019 | 0.000 | 0.000 | 0.000 | 0.000 | 0.000 | 0.000 | 0.013 |
| | | 2-bit | 1.9967 | 0.083 | 0.175 | 0.000 | 0.000 | 0.000 | 0.000 | 0.000 | 0.000 | 0.018 | 0.000 | 0.000 | 0.022 |
| | | 3-bit | 1.7822 | 0.125 | 0.250 | 0.755 | 0.977 | 0.170 | 0.787 | 0.939 | 0.365 | 0.214 | 0.000 | 0.000 | 0.409 |
| | | 4-bit | 0.7930 | 0.125 | 0.150 | 0.811 | 0.886 | 0.642 | 0.936 | 0.980 | 0.192 | 0.179 | 0.041 | 0.000 | 0.446 |
| | | 5-bit | 0.3301 | 0.000 | 0.050 | 0.377 | 0.636 | 0.509 | 0.809 | 0.878 | 0.096 | 0.018 | 0.041 | 0.038 | 0.309 |
| | Median Smoothing | 2x2 | 1.1296 | 0.188 | **0.550** | **0.981** | **1.000** | 0.717 | 0.979 | **1.000** | **0.981** | **1.000** | **0.837** | **0.885** | 0.836 |
| | | 3x3 | 1.9431 | 0.042 | 0.250 | 0.660 | 0.932 | 0.038 | 0.681 | 0.918 | 0.750 | 0.929 | 0.041 | 0.077 | 0.486 |
| | Non-local Mean | 11-3-2 | 0.2770 | 0.125 | 0.400 | 0.830 | 0.955 | 0.717 | 0.915 | 0.939 | 0.077 | 0.054 | 0.265 | 0.154 | 0.484 |
| | | 11-3-4 | 0.7537 | 0.167 | 0.525 | 0.868 | 0.977 | 0.679 | 0.936 | **1.000** | 0.250 | 0.232 | 0.245 | 0.269 | 0.551 |
| | | 13-3-2 | 0.2910 | 0.125 | 0.375 | 0.849 | 0.977 | 0.717 | 0.915 | 0.939 | 0.077 | 0.054 | 0.286 | 0.173 | 0.490 |
| | | 13-3-4 | 0.8290 | 0.167 | 0.525 | 0.887 | 0.977 | 0.642 | 0.936 | **1.000** | 0.269 | 0.232 | 0.224 | 0.250 | 0.547 |
| | Best Attack-Specific Single Squeezer | | - | 0.188 | **0.550** | **0.981** | **1.000** | 0.717 | 0.979 | **1.000** | **0.981** | **1.000** | **0.837** | **0.885** | - |
| | Best Joint Detection (5-bit, 2x2, 13-3-2) | | 1.1402 | **0.208** | **0.550** | **0.981** | **1.000** | **0.774** | **1.000** | **1.000** | **0.981** | **1.000** | **0.837** | **0.885** | **0.845** |
| ImageNet | Bit Depth | 1-bit | 1.9942 | 0.151 | 0.444 | 0.042 | 0.021 | 0.048 | 0.064 | 0.000 | 0.000 | 0.000 | - | - | 0.083 |
| | | 2-bit | 1.9512 | 0.132 | 0.511 | 0.500 | 0.354 | 0.286 | 0.170 | 0.306 | 0.218 | 0.191 | - | - | 0.293 |
| | | 3-bit | 1.4417 | 0.132 | 0.556 | **0.979** | **1.000** | 0.476 | 0.787 | **1.000** | 0.836 | **1.000** | - | - | 0.751 |
| | | 4-bit | 0.7996 | 0.038 | 0.089 | 0.813 | **1.000** | 0.381 | 0.915 | **1.000** | 0.727 | **1.000** | - | - | 0.664 |
| | | 5-bit | 0.3528 | 0.057 | 0.022 | 0.688 | 0.958 | 0.310 | **0.957** | **1.000** | 0.473 | **1.000** | - | - | 0.606 |
| | Median Smoothing | 2x2 | 1.1472 | 0.358 | 0.422 | 0.958 | **1.000** | 0.714 | 0.894 | **1.000** | **0.982** | **1.000** | - | - | 0.816 |
| | | 3x3 | 1.6615 | 0.264 | 0.444 | 0.917 | 0.979 | 0.500 | 0.723 | 0.980 | 0.909 | **1.000** | - | - | 0.749 |
| | Non-local Mean | 11-3-2 | 0.7107 | 0.113 | 0.156 | 0.813 | 0.979 | 0.357 | 0.936 | 0.980 | 0.418 | 0.830 | - | - | 0.618 |
| | | 11-3-4 | 1.0387 | 0.208 | 0.467 | 0.958 | **1.000** | 0.548 | 0.936 | **1.000** | 0.673 | 0.957 | - | - | 0.747 |
| | | 13-3-2 | 0.7535 | 0.113 | 0.156 | 0.813 | 0.979 | 0.357 | 0.936 | 0.980 | 0.418 | 0.851 | - | - | 0.620 |
| | | 13-3-4 | 1.0504 | 0.226 | 0.444 | 0.958 | **1.000** | 0.548 | 0.936 | **1.000** | 0.709 | 0.957 | - | - | 0.751 |
| | Best Attack-Specific Single Squeezer | | - | 0.358 | 0.556 | **0.979** | **1.000** | 0.714 | **0.957** | **1.000** | **0.982** | **1.000** | - | - | - |
| | Best Joint Detection (5-bit, 2x2, 11-3-4) | | 1.2128 | **0.434** | **0.644** | **0.979** | **1.000** | **0.786** | 0.915 | **1.000** | **0.982** | **1.000** | - | - | **0.859** |

**Training.** The training phase of our detector is simply selecting an optimal threshold of $score^{joint}$. One typical practice is to find the one that maximizes the training accuracy. Since the actual expected distribution of samples is not balanced and mostly benign, however, a detector with high accuracy but a high false positve rate would be useless for many security-sensitive tasks. Therefore, we instead select a threshold that targets a false positive rate below 5% by choosing a threshold that is exceeded by no more than 5% of legitimate samples. Note that the training threshold is set using only the legitimate examples, so does not depend on the adversarial examples. Consequently, our method is inexpensive in the training phase compared with other methods like *sample statistics* or *training a detector* as we introduce in Section 2-D, but could be more expensive than *training a detector* during inference due to the squeezing operations and the multiple inputs.

**Validation.** Next, we use the chosen threshold value to measure the detection rate on three groups: successful adversarial examples (SAEs), failed adversarial examples (FAEs), and legitimate examples (for false positive rate). Except when noted explicitly, "detection rate" means the detection rate on successful adversarial examples. It is important to distinguish failed adversarial examples from legitimate examples here since detecting failed adversarial examples is useful for detecting attacks early, whereas an alarm on a legitimate example is always undesirable and is counted as a false positive.

### C. Results

Table 4 shows the detection rates for successful adversarial examples for each attack method with a variety of configurations. For each dataset, we first list the detection rate for several detectors built upon single squeezers. For each squeezing method, we tried several parameters and compare the performance for each dataset. The "Best Attack-Specific Single Squeezer" row gives the detection rate for the best single squeezer against a particular attack. This represents the (unrealistically optimistic) case where the model operator knows the attack type and selects a single squeezer for detection that may be different for each attack. Below this, we show the best result of joint detection (to be discussed later) with multiple squeezers where the same configuration must be used for all attacks.

The best bit depth reduction for MNIST is squeezing the color bits to one, which achieves at least 97.87% detection for all the $L_\infty$ and $L_2$ attacks and 100% detection rate for seven of the attacks. It is not as effective on CW$_0$ attacks, however, since these attacks are making large changes to a small number of pixels. On the contrary, the $3 \times 3$ median smoothing is the most effective on detecting the $L_0$ attacks with detection rates above 88%. This matches the observation from Table 3 that they have different strengths for improving the model accuracy. For MNIST, there is at least one squeezer that provides good (> 91%) detection results for all of the attacks.

For CIFAR-10, $2 \times 2$ median smoothing is the best single squeezer for detecting every attack except DeepFool, which is best detected by non-local means. This is consistent with the robustness results in Table 3. For the ImageNet dataset, we

TABLE 5: Summary results for the best joint detectors.

| Dataset | Detector | Threshold | Detection (SAEs) | Detection (FAEs) | FPR | ROC-AUC (Excluding FAEs) |
|---|---|---|---|---|---|---|
| MNIST | Bit Depth (1-bit), Median (2×2) | 0.0029 | 98.15% | 20.00% | 3.98% | 99.44% |
| CIFAR-10 | Bit Depth (5-bit), Median (2×2), Non-local Means (13-3-2) | 1.1402 | 84.53% | 22.22% | 4.93% | 95.74% |
| ImageNet | Bit Depth (5-bit), Median (2×2), Non-local Means (11-3-4) | 1.2128 | 85.94% | 25.00% | 8.33% | 94.24% |

SAE: successful adversarial example. FAE: failed adversarial example.

TABLE 6: Comparison with MagNet.

| Dataset | Method | AEs | SAEs |
|---|---|---|---|
| MNIST | Feature Squeezing | 69.08% | 78.75% |
| MNIST | MagNet | **91.77%** | **95.61%** |
| CIFAR-10 | Feature Squeezing | **60.87%** | **61.88%** |
| CIFAR-10 | MagNet | 50.36% | 50.46% |

False positive rate for both is 0.40% on MNIST, 1.28% on CIFAR-10.

find several different squeezers are similarly effective on each attack type. For example, the $CW_2$-LL attack can be detected 100% of the time with several bit depth filters, $2 \times 2$ median smoothing, and some of the non-local mean filters.

The third column in the table gives the distance threshold setting needed to satisfy the maximum false positive rate of 5% in the training dataset. These threshold values provide some insight into how well a particular squeezer distinguishes between adversarial and legitimate examples. For the binary filter on MNIST, a tiny threshold value of 0.0005 was sufficient to produce a false positive rate below 5%, which means the squeezing has negligible impact on the legitimate examples: 95% of the legitimate examples have the $L_1$-based distance score below 0.0005. On the other hand, the best median smoothing filter ($2 \times 2$) on MNIST needs a larger threshold value 0.0029 to achieve a similar false positive rate, which means it is slightly more destructive than the binary filter on the legitimate examples. The more aggressive median smoothing with $3 \times 3$ window results in an even higher threshold 0.039, because the legitimate examples could get over-squeezed to the target classifier. A lower threshold is always preferred for detection since it means the detector is more sensitive to adversarial examples.

For some of the attacks, none of the feature squeezing methods work well enough for the color datasets. The worst cases, surprisingly, are for FGSM and BIM, two of the earlier adversarial methods. The best single-squeezer-detection only recognizes 18.75% of the successful FGSM examples and 55% of BIM examples on the CIFAR-10 dataset, while the detection rates are 35.85% and 55.56% on ImageNet. We suspect the reason the tested squeezers are less effective against these attacks is because they make larger perturbations than the more advanced attacks (especially the CW attacks), and the feature squeezers we use are well suited to mitigating small perturbations. Understanding why these detection rates are so much lower than the others, and developing feature squeezing methods that work well against these attacks is an important avenue for future research.

**Joint-Detection with Multiple Squeezers.** By comparing the last two rows of each dataset in Table 4, we see that joint-detection often outperforms the best detector with a single squeezer. For example, the best single-squeezer-detection detects 97.87% of the $CW_2$-Next examples for CIFAR-10, while joint detection detects 100%.

The main reason to use multiple squeezers, however, is because this is necessary to detect unknown attacks. Since the model operator is unlikely to know what attack adversaries may use, it is important to be able to set up the detection system to work well against any attack. For each data set, we try several combinations of the three squeezers with different parameters and find out the configuration that has the best detection results across all the adversarial methods (shown as the "Best Joint Detection" in Table 4, and summarized in Table 5). For MNIST, the best combination was the 1-bit depth squeezer with $2 \times 2$ median smoothing (98.15% detection), combining the best parameters for each type of squeezer. For the color image datasets, different combinations were found to outperform combining the best squeezers of each type. The best joint detection configuration for ImageNet (85.94% detection) includes the 5-bit depth squeezer, even though the 3-bit depth squeezer was better as a single squeezer.

Since the joint detector needs to maintain the 5% false positive rate requirement, it has a higher threshold than the individual squeezers. This means in some cases its detection rate for a particular attack will be worse than the best single squeezer achieves. However, comparing the "Best Attack-Specific Single Squeezer" and "Best Joint Detection" rows in Table 4 reveals that the joint detection is usually competitive with the best single squeezers over all the attacks. For MNIST, the biggest drop is for detection rate for $CW_0$ (LL) attacks drops from 100% to 93%; for CIFAR-10, the joint squeezer always outperforms the best single squeezer; for ImageNet, the detection rate drops for $CW_2$ (Next) (95% to 91%). For simplicity, we use a single threshold across all of the squeezers in a joint detector; we expect there are better ways to combine multiple squeezers that would use different thresholds for each of the squeezers to avoid this detection reduction, and plan to study this in future work.

We report ROC-AUC scores in Table 5 excluding the failed adversarial examples from consideration, since it is not clear what the correct output should be for a failed adversarial example. Our joint-detector achieves around 95% ROC-AUC score on CIFAR-10 and ImageNet. The ROC-AUC of the detector is as high as 99.44% for MNIST. The false positive rates on legitimate examples are all near 5%, which is expected considering how we select a threshold value in the training phase. The detection rate for the best configuration on successful adversarial examples exceeds 98% for MNIST using a 1-bit filter and a 2×2 median filter and near 85% for the other two datasets using a combination of three types feature squeezing methods with different parameters. The detection rates for failed adversarial examples are much lower than those for successful adversarial examples, but much higher than the false positive rate for legitimate examples. This is unsurprising since FAEs are attempted adversarial examples, but since they are not successful the prediction outputs for the unsqueezed and squeezed inputs are more similar.

We compare our results with MagNet [24] in Table 6. We configured the MagNet detectors on two datasets following the description in their paper and reported the detection performance with our target models and the detection dataset. In order to fairly compare the detection rates, we adjusted the threshold values of our detectors accordingly on the two datasets to produce the false positive rates matching the MagNet results: 0.40% for MNIST and 1.28% for CIFAR-10. MagNet achieves higher detection rates on MNIST (91.77% over 69.08%), while our method outperformed on CIFAR-10 (60.87% over 50.36%). The detection rates excluding failed adversarial examples were similar. MagNet's detection rates on our adversarial examples for MNIST are impressive, and superior to what the best feature squeezing configuration achieves. However, this advantage does not apply to CIFAR-10. Further, it is more expensive to use MagNet because it requires training an autoencoder on a whole dataset. In addition, the MagNet pipeline is end-to-end differentiable, making it vulnerable to trivial white-box adversary.

### D. Adversarial Adaptation

So far, we have only considered static adversaries who do not adapt to attack our feature squeezing method directly. Now, we consider adaptive adversaries who have full knowledge of the defense. To be successful against our detection framework, an adversary needs to find an input where the original classifier produces the wrong output and the $L_1$ score between the model's predictions on squeezed and original inputs is below the detection threshold. This is a much harder problem than just finding an adversarial example, as is supported by our experimental results.

He et al. [13] recently proposed an adaptive attack which can successfully find adversarial examples that defeat one configuration of a feature squeezing defense.[3] The approach finds adversarial examples that both confuse the original model and have a $score^{joint}$ lower than a pre-selected threshold for squeezed inputs. Their approach adapts the $CW_2$ attack by adding a penalty term for the $L_1$ prediction distance. It requires that all the feature squeezing operators are differentiable so that it is possible to compute the gradient of the loss function in the optimization process. For the non-differentiable feature squeezers such as the bit depth reduction, their approach requires restarting the algorithm several times with random initialization and hoping it finds an example that is resilient against the non-differentiable squeezers. This means the attack is non-deterministic and more time-consuming in face of non-diffrentiable components [13]. The attack takes roughly 20 seconds on each MNIST sample, which is around 60 times slower than the original $CW_2$ attack.

We only evaluate their adaptive attack on the MNIST dataset, because we currently don't have a Tensorflow implementation of the non-local means filter used on CIFAR-10 and ImageNet.[4] We adapted their attacking code to our MNIST model and the joint-detector we report in Table 5.

In addition to the untargeted attack He et al. tested, we also tested the two targeted attacks mentioned earlier in Section 4. We used the same parameters as He et al. [13] in generating the adaptive adversarial examples except that we change the target threshold value to 0.0029 that is used by the detector in this paper.[5] The seed examples are the same 100 images used in Section 4. With the exact threshold we use in this paper as the target value, the success rate of the untargeted attack is 68% (compared with 100% success rate in He et al.'s report [13]). The success rates of the targeted attacks are even lower: 44% for the targeted-next attack and 24% for the targeted-ll attack. We believe these lower adversarial success rates are due to using a different MNIST model (representative of a state-of-the-art model) and our detector has a much smaller threshold value (0.0029) compared to the one they used (over 0.3076).

However, the success rates reported in the adaptive attack do not provide a useful measure of robustness because the discussed adaptive adversary is allowed to modify any pixel arbitrarily. Given unlimited perturbation bounds, one could always force a model to misclassify an example into any class, but often by producing images that are unrecognizable or obviously suspicious to humans. Figure 6 shows typical examples produced by the adaptive attack. In order to understand the robustness of our detector in a more realistic setting, we clipped the perturbations with different $\epsilon$ values ranging from 0 to 1. Figure 7 shows the adversary success rates as the adversary's strength $\epsilon$ increases to the right. At 0.3, the $\epsilon$ value considered a reasonable limit on what is a permissible adversarial example by many papers [21], the adversary's success rate is 0% for both of the target attacks and 1% for the untargeted attack.

**Countermeasures.** One obvious strategy to combat adaptive adversaries is to introduce randomness in the squeezing method. This is very different from attempts to obfuscate models, which have been shown vulnerable to transfer attacks. Instead, we can use cryptographic randomness to make the deployed framework unpredictable, since the adversary's search requires knowledge of the exact squeezing operation. The defender has many opportunities to use randomness in selecting squeezing parameters. For example, instead of using a fixed 0.5 threshold for the 1-bit filter, using $0.5 \pm rand(0.1)$ (which could be done with a different random value for each pixel), or selecting random regions for the median smoothing instead of a fixed $2 \times 2$ region).

We conducted an experiment in which the cutoff value of the binary filter follows a normal distribution with 0.5 mean and standard deviation 0.0625. The success rates (with no limit on perturbation magnitude) decrease from $(68\%, 44\%, 24\%)$ to $(17.0 \pm 2.4\%, 16.2 \pm 2.6\%, 7.6 \pm 2.2\%)$ (respectively for untargeted, targeted to next class, and targeted to least likely; measured over 10 executions).

---

[3]This work was done following initial public reports on the work in this paper [42, 43]; we shared details of our approach and code with the authors of [13], and much appreciate their sharing their implementation with us to enable the experiments reported here.

[4]He et al. reported results for CIFAR-10 with only bit depth reduction and median smoothing [13]. These results were similar to the results they reported on MNIST.

[5]He et al.'s target detector was based on an earlier version of this work that used 0.3076 as threshold and uses a slightly different max function in combining multiple squeezers.

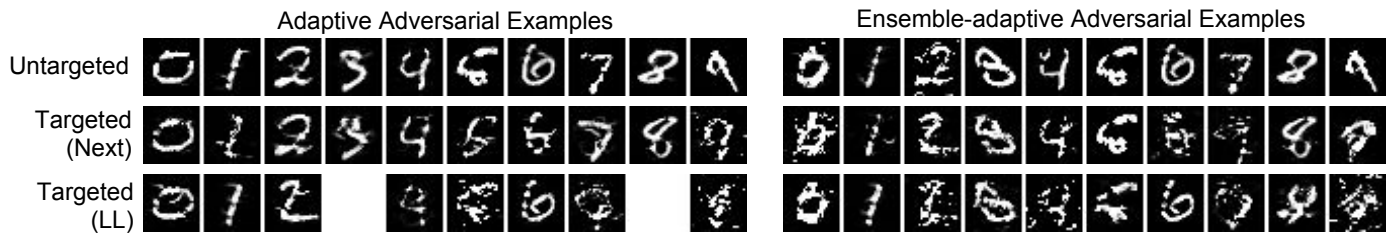

Fig. 6: **Adversarial examples generated by the adaptive adversary.** The images are randomly sampled from the successful adversarial examples generated by the adaptive adversarial methods. No successful adversarial examples were found for Targeted (LL) 3 or 8. The average $L_2$ norms of the successful adversarial examples are respectively 2.80, 4.14, 4.67 for the untargeted, targeted (next) and targeted (ll) examples; while the corresponding values are 3.63, 5.48, 5.76 for the ensemble-adaptive adversarial examples. The average $L_\infty$ norm are 0.79, 0.89, 0.88 for the adaptive adversarial examples; while the corresponding values are 0.89, 0.95 and 0.96 for the ensemble-adaptive adversarial examples.

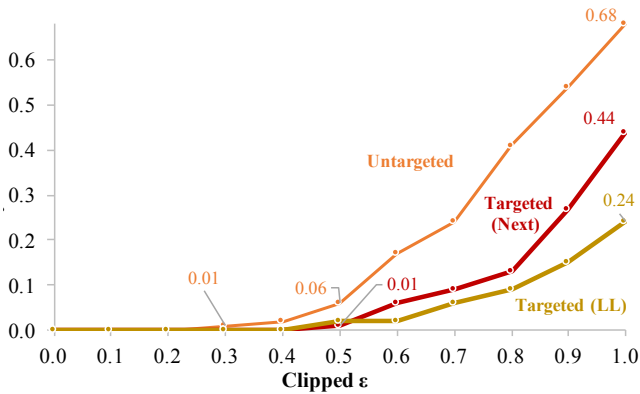

Fig. 7: Adaptive adversary success rates.

An adversary may attempt to adapt to the randomness by attacking an ensemble of random squeezers. We consider an ensemble-adaptive adversary that considers three thresholds of the binary filter together: 0.4, 0.5 and 0.6. The success rates increase to $(46.5 \pm 2.1\%, 34.5 \pm 3.5\%, 28.5 \pm 2.1\%)$ measured over 10 executions. However, the perturbations became even larger, resulting in many unrecognizable and suspicious-looking images (shown in the right part of Figure 6).

## 6. Conclusion

The effectiveness of feature squeezing seems surprising since it is so simple and inexpensive compared to other proposed defenses. Developing a theory of adversarial examples remains an illusive goal, but our intuition is that the effectiveness of squeezing stems from how it reduces the search space of possible perturbations available to an adversary.

Although we have so far only experimented with image classification models, the feature-squeezing approach could be used in many domains where deep learning is used. For example, Carlini et al. demonstrated that lowering the sampling rate helps to defend against the adversarial voice commands [5]. Hosseini et al. proposed that correcting the spelling on inputs before they are provided to a character-based toxic text detection system can defend against adversarial examples [15].

As discussed in Section 5-D, feature squeezing is not immune to adversarial adaptation, but it substantially changes the challenge an adversary faces. Our general detection framework opens a new research direction in defending against adversarial examples and understanding the limits of deep neural networks in adversarial contexts.

## Availability

The models and our implementations of the attacks, defenses, and tests, are available in the EvadeML-Zoo open source toolkit (https://evadeML.org/zoo).

## Acknowledgments

This work was partially supported by grants from the National Science Foundation (SaTC, #1619098), Intel Corporation, and cloud computing research credits from Amazon and Microsoft. The authors thank Nicolas Papernot for tremendous help improving the presentation of the paper, and for providing Cleverhans to the research community. The authors thank Warren He for sharing the code which enabled our experiment with the adaptive adversary, Dongyu Meng for sharing MagNet and the Madry Lab for sharing PGD. We thank Noah Kim and Andrew Norton for contributions to EvadeML-Zoo.